# Inverse Design of Optimal Stern Shape with Convolutional Neural Network-based Pressure Distribution

Sang-jin Oh (Pusan National University), Ju Young Kang (Pusan National University),
Kyungryeong Pak (Samsung Heavy Industries co., Ltd.), Heejung Kim (Samsung Heavy Industries co., Ltd.), and Sung-chul Shin (Pusan National University)

## Abstract

Hull form designing is an iterative process wherein the performance of the hull form needs to be checked via computational fluid dynamics calculations or model experiments. The stern shape has to undergo a process wherein the hull form variations from the pressure distribution analysis results are repeated until the resistance and propulsion efficiency meet the design requirements. In this study, the designer designed a pressure distribution that meets the design requirements; this paper proposes an inverse design algorithm that estimates the stern shape using deep learning. A convolutional neural network was used to extract the features of the pressure distribution expressed as a contour, whereas a multi-task learning model was used to estimate various sections of the stern shape. We estimated the stern shape indirectly by estimating the control point of the B-spline and comparing the actual and converted offsets for each section; the performance was verified, and an inverse design is proposed herein

## 1. Introduction

Hull form is the outer shape of the hull streamlined to satisfy the requirements of the ship owner such as those related to deadweight and ship speed. A fundamental task in hull form designing is to develop a hull form that performs well in various aspects including resistance, propulsion, and maneuvers [1]. Specifically, in the case of stern shape, the flow field formed in the hull changes according to the stern, and it is related to propulsion as well as resistance in the hull [2]. Therefore, it is important to design the hull form such that the distribution of pressure entering the stern can achieve the optimal performance possible. Hull form is extracted through its variations [3-6] according to the requirements, and the performance is evaluated using model testing and CFD [4, 7-9]. This process is carried out until the constraint is met [10], otherwise the hull form variation parameters are modified again to create a new hull form [11]. Although the number of iterations and design time were reduced through the practical engineering application of the SBD method and the application of the approximation method [12] to reduce the computation time of the CFD operation, many researchers and design engineers started designing with unsatisfactory configurations on a trial-and-error basis, trying two or three different configurations before eventually selecting the best one [13]. Because hull form design finds the initial design point based on the designer's experience and model test database, experienced designers make efficient designs to reduce repetitive tasks and deliver satisfactory performance, but relatively inexperienced designers find appropriate design points is difficult failure to properly identify the initial design points increases uncertainty about current and future costs [14]. We propose a hull form that achieves maximum hydrodynamic performance within constraints through an inverse design process to reduce iterations.

In order to ascertain the optimal hull shape at the initial design point, there is a direct method [15] that determines the one with the maximum efficiency among various hull forms. In contrast, there is an inverse method which employs previous data to find a hull form for a new optimization object. [16] use an inverse method to specify the desired pressure distribution and inverse design of the airfoil shape through an optimization process. While the hull form optimization method classified as the direct method requires complex algorithms and numerous forward calculations, the inverse method requires fewer calculations and solves the classic problem of determining the hydrodynamic characteristics that represent the desired pressure distribution [17]. In the design methods using inverse design [15, 17], the dimension of the algorithm changes depending on the output shape. In [18], the control points were updated as parameters in Non-Uniform Rational B-spline (NURBS), with values smaller than the initial wave-making resistance and specifically, a total of seven parameters were performed in the inverse design method of the overall algorithm by designing a stem shape including a bulbous bow. To handle pressure distributions with spatial information, it was necessary to process more complex two-dimensional optimization targets. To solve this problem, [19] used CNNs to select two-dimensional pressure distribution ($C_p$) images as the target of optimization and inverse design of airfoil shape by optimizing algorithms. Using a pressure distribution image of 144×144 as an input, it is designed to predict y-axis points at 120 fixed x-axis points. Although two-dimensional optimization targets have been used to improve the accuracy of shape prediction, application to the three-dimensional stern shape requires more prediction than the airfoil with a two-dimensional shape, also requiring inverse design methods to fit the stern shape.

CNNs used in [19] are being actively researched as a method of analyzing and processing data as a deep learning field, surpassing the performance of previous machine learning-based algorithms [20]. CNNs have higher accuracy than humans and exhibit excellent performance especially in terms of image processing. Therefore, it has been used in various fields including image classification and object detection fields [21]. The basic function of CNN is to extract features for images from a convolutional layer, and to find and classify features using an existing MLP-type layer known as a fully connected layer. CNNs are fast and accurate as they can solve both functions of image feature extraction and classification within one algorithm [22] Because the output of the MLP layer corresponding to the task is in the form of a one-dimensional array [23-25], it must be converted into a one-dimensional array in order to estimate the stern shape composed of three-dimensions. The converted output loses spatial information, and there is a risk of overfitting / underfitting the estimation model by increasing the number of output nodes in one task. In [26], overfitting was prevented by reducing the number of outputs that need to be processed by the task by sharing the hidden layer between tasks and having a separate output layer for each task. This learning method is called multi-task learning, and it is used for estimation utilizing images as input [27] or for multi-class classification [26]. Based on this concept, if the *x*-axis corresponding to the longitudinal direction of the hull is divided into sections and each section is set as a task, spatial information is expected to be partially reflected.

As shown in Fig. 1, we propose an inverse design algorithm using CNNs to estimate the stern shape that satisfies the hydrodynamic performance set by the designer. To express the stern shape divided into sections, the task of CNN is composed of multi-task learning, making it an algorithm suitable for the inverse design of the stern shape. In the experiment, offset data obtained by hull form variation of KRISO VLCC (KVLCC) and image data analyzed through CFD were used. In addition, to reduce the number of output nodes, the data was reprocessed by converting the offset into a specific value in the data preprocessing phase. Finally, for the performance test of the algorithm, the basic single-task learning model and the multi-task learning model were compared, and the performance change according to the contour in the CFD image was also confirmed.

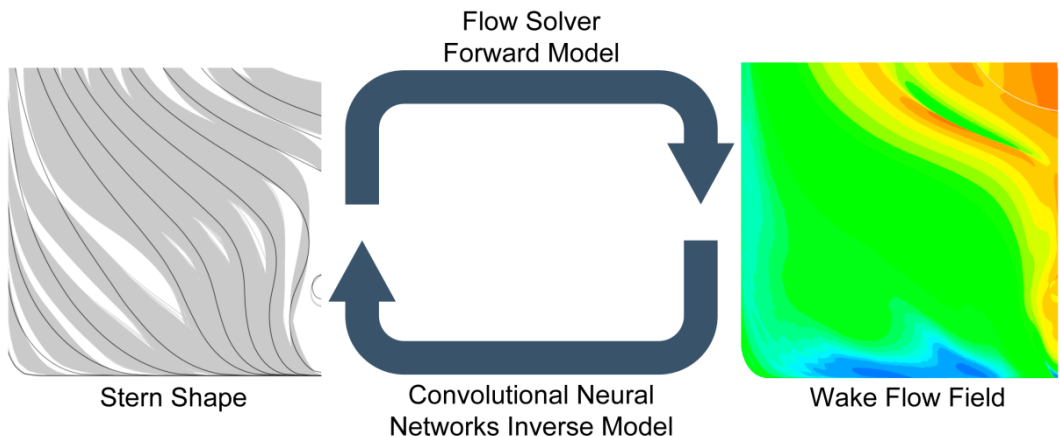

Fig. 1. Inverse design algorithm using CNNs to estimate the stern shape.

## 2. Methodology

Machine learning is largely divided into supervised learning and unsupervised learning. The scope of the experiment is limited to supervised machine learning applied to regression problems, where an algorithm is trained using labeled inputs to extract features from the data. The algorithm to be used in this study was selected as CNNs, and the basic configuration is as follows.

The data of multiple neurons connected to each node are received, synthesized and processed, and then the processed data are delivered to the next node. These connected neurons correspond to weights, which are combined with the input node, after which a bias is added to them. Subsequently, the level of expression of the next node is determined through the specified activation function. These nodes are accumulated and form one layer between input and output layers, and deliver the features entering through the input layer to the next layer. This layer located between the input and output layers is called a hidden layer, and a stack of multiple hidden layers is called the MLP, with a deep form of many hidden layers known as deep neural networks (DNNs) [28]. CNNs have been changed into a form that can reflect spatial information using the basic concept of DNNs.

The most characteristic layer constructed to enable the network to handle three-dimensional data and extract the features of images is the convolutional layer. The weights between input and output layers are shared as filters. In the case of DNNs, there was a parameter between nodes and there were N×N parameters in total. However, the convolutional layers are shared and operated through one filter and efficient calculation is possible because the number of parameters correspond according to the filter size [22]. The width and height were calculated at regular intervals, and information about the location is represented as a feature value. Hence, image recognition is possible because of the information about the location. A fixed interval is expressed as a stride, and this determines how much the filter should be moved. As shown in Fig. 2, the outputs were accumulated and the filter and input values are calculated starting from the top left corner in the input feature map. The channels represent color or depth in three-dimensional data and have filters of the specified number. After operating with the corresponding filter, all channels are added together to produce a single output feature map. If there are multiple filters, multiple output feature maps are produced. Before moving on to the next layer in the output, the degree of expression is calculated and determined through the activation function.

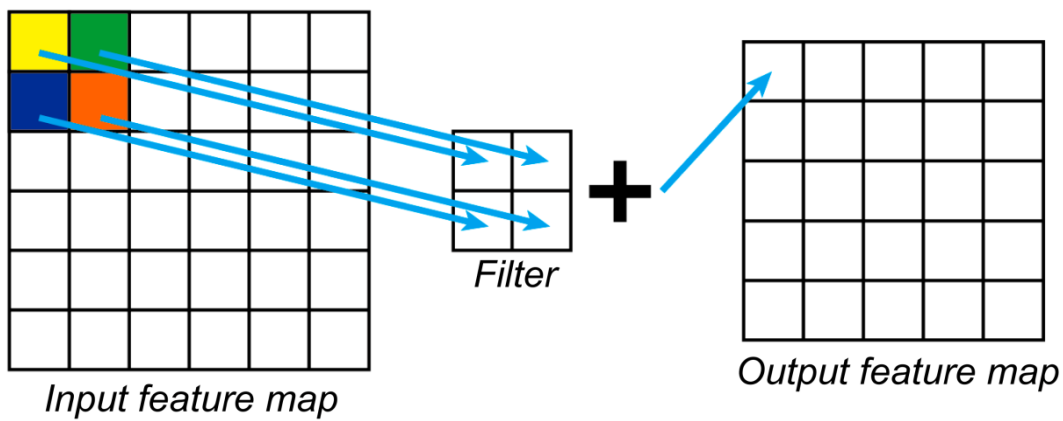

Fig. 2. Convolutional layer: As the filter moves by 1 Stride, the features of the input feature map are calculated. The input function map and the 2×2 filters are multiplied elementwise, then summed to give one output feature value.

The pooling layer generally follows the convolutional layer and functions to reduce the size of the feature map and extract representative values. The convolutional layer requires many calculations and a large memory because it has filters and weights which need to be trained, however, the pooling layer does not need many calculations because it calculates the maximum or average within the filter size, which does not need to be trained. Among the pooling methods, average pooling, which obtains the average, reduces the value of the feature map produced through the activation function and is mainly used in the image of average. Max pooling, which calculates the maximum value, can extract information about an image by selecting outstanding features and shows higher efficiency than average pooling in images without specific objects [29].

The fully connected layer functions to classify or regress features extracted from the convolutional layer and the pooling layer and has the same form as the MLP where all the weights between nodes are connected. When a network is set up as a final output of the fully connected layer, the output formed through data analysis is received as one-dimensional data.

Most CNN models perform one task through a feature extractor (convolutional layer, pooling layer) and regressor (fully connected layer). There are three methods for estimating $n$ information in one image.

- Method 1: Feature extractor 1, Regressor $S$ – create $S$ models.
- Method 2: Feature extractor 1, Regressor 1 - create 1 model.
- Method 3: Feature extractor 1, Regressor $S$ – create 1 model.

In the case of $S$ models, there is an advantage in that a model optimized for each set of information can be generated, however, image features are duplicated for each model, so the number of parameters to be calculated increases and the correlation between information disappears. In a deep learning algorithm, in order to reflect not only information but also relationships with each other, it must be estimated from one model. When all information is estimated in one regressor, the model becomes simple and can reflect the relationship, however, the information is not differentiated and the model is trained on average. Furthermore, as the output node increases, the number of parameters of the fully connected layer increases, and there is a risk of overfitting / underfitting. A model that can learn by dividing $S$ information in one model can be integrated into one model by sharing a feature extractor, and $S$ regressors learn about each information to solve the problems of Method 1 and Method 2. We intend to estimate various information of the stern shape using multi-task learning, a method for learning $S$ regressors.

## 3. Data Preprocessing

Data processing is the most important part in deep learning, and increasing the weight of data processing and allowing each weight to be properly trained is so vital that it takes precedence over a network architecture or a combination of algorithm options [30]. Inverse design is used to estimate the stern shape according to the hydrodynamic performance, and in this case, the hydrodynamic performance set by the designer refers to the image from the CFD analysis result. To reflect a wide range of CFD analysis results in the algorithm, various stern shape and CFD analysis results were used as experimental data. Based on KVLCC, 1,002 were generated through hull form variation, and images were extracted by CFD calculation using STARCCM+. As shown in Fig. 3, the contour levels, which represent the gaps of flow fields in STARCCM+ were divided into 25 and 35 to examine the effects of contour levels on learning in more detail. In addition, the effects of images including lines were also examined by composing data to distinguish contour levels.

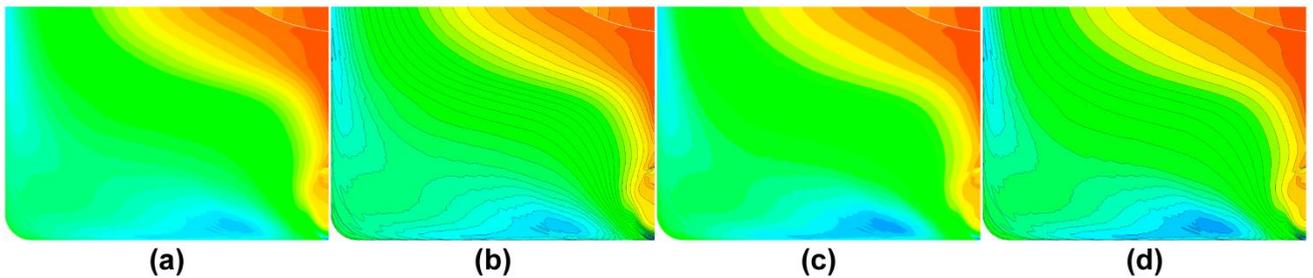

Fig. 3. Experimental images: (a) contour level 35 without lines (b) contour level 35 with lines
(c) contour level 25 without lines (d) contour level 25 with lines

The offset of the stern shape indicates the three-dimensional shape of the hull, with the length, width and depth indicated by the x-axis, y-axis and z-axis respectively. As shown in Fig. 4, sections are divided differently depending on the ship in the x-axis direction, and in the experiment, the stern of the KVLCC up to the midship section was divided into 13 frames (-5.5, 0, 8, 16, 24, 32, 40, 48, 64, 80, 96, 112, 160 – referring to the distance from the after perpendicular in meters). Here, the section corresponding to 8 m has a transom and a stern bulb, so different section index numbers were assigned to distinguish them. In each section, the number of coordinate points is displayed differently depending on the size or setting of the curve. There are 14 section data used for stern shape estimation, and each section consists of y-axis and z-axis coordinate values corresponding to the half width of the ship. Coordinate points composed of pairs of originals before conversion into 14 points are extracted in the process of hull form variation. In this process, they are automatically extracted according to the program settings or can be set arbitrarily by the designer. In this study, to prevent excessive node creation and to process data quickly, 50 per section were fixed and extracted. The half-width of the width is 29 m, which is the maximum value of the y-coordinate, and the depth is 21 m, which appears as the maximum value of the z-coordinate.

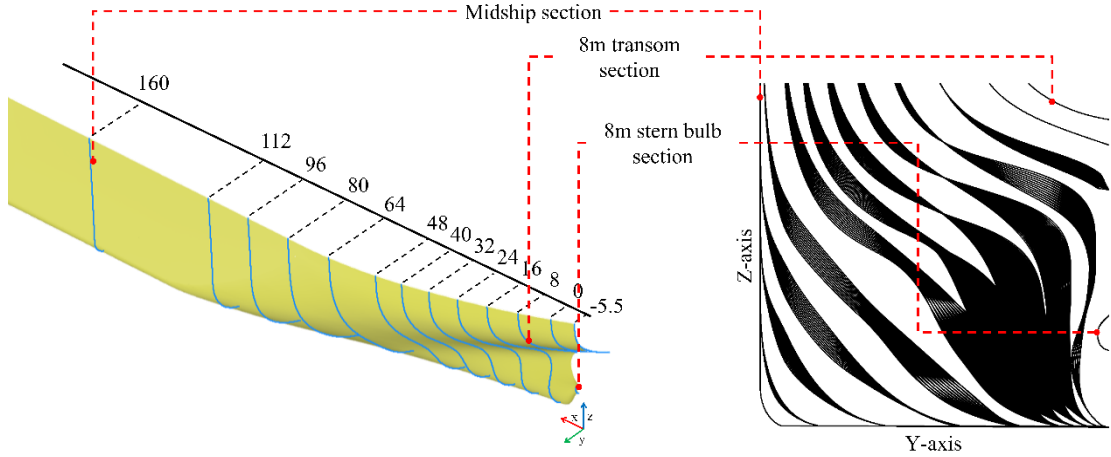

Fig. 4. 1,002 hull form by dividing the stern of KVLCC into 13 frames to the midship.

The stern shape contains shape information for each section. It is difficult to define a certain number of sections because the number of offsets is different for each section according to the hull form. Because the output of CNNs should have a fixed number owing to network characteristics, the coordinate points should be set to a specific number. If the output is different, the number of learned parameters is different, and it cannot be learned with integrated data while changing the structure. This is similar to having a different structure for each data. Although we fixed the original coordinates to 50 and extracted them, the number of parameters that need to be learned compared to the number of data increases to be used as the output of the algorithm. Therefore, two methods were used to solve this problem.

- Removal of straight sections.
- Calculate control points for B-spline curves.

The ship is fixed in width and height according to the principal dimensions and has parallel sections according to Flat of Side (FOS) and Flat of Bottom (FOB). When the hull form is expressed in 2-D, it corresponds to a straight section which is not required for learning because it is a predetermined value. Therefore, the data were constructed after removing the straight section before training. Additionally, because the straight section causes noise in the B-spline later, it has a positive effect on the B-spline. If straight line coordinate points are removed from the original data, the number of coordinate points vary for each section.

The data were reprocessed using B-spline to express a hull form with a constant number of coordinate points. The B-spline curve equation is used for curve fitting of multi-dimensional data and is expressed as a hull form combination of control points and B-spline basis functions, with its arrangement determined by the order of the B-spline basis functions. In most cases, a quadratic B-spline ($k = 3$) curve as good as a cubic B-spline ($k = 4$) curve is generated [31]. In this study, a quadratic B-spline curve with a uniform open knot was used for efficiency. The knot vector of the B-spline curve with $(n + 1)$ control points approximating the given $(m + 1)$ data $P_i$ is expressed as follows:

$$0.0 = t_0 = t_1 = \cdots = t_{k-1} < t_k < \cdots < t_n < t_{n+1} = \cdots = t_{n+k} = 1.0 \qquad (1)$$

The parameters corresponding to $(m + 1)$ data were determined by the chord length approximation method as in Eq.

(2).

$$u_i = \begin{cases} 0.0, & i = 0, \\ \sum_{j=0}^{i-1}|P_{j+1} - P_j| \Big/ \sum_{j=0}^{m-1}|P_{j+1} - P_j|, & i = 1, \ldots, m-1, \\ 1.0, & i = m. \end{cases} \quad (2)$$

The relationship between the given data $P_i$ and the control point is expressed in Eq. (3) by the B-spline curve equation.

$$P_i(u_i) = \sum_{j=0}^{n} Q_j N_j^k(u_i) \quad (3)$$

Here, $N_j^k$ is related to Eq. (1), and is defined as a basic function of a quadratic B-spline as follows:

$$N_j^k(u) = \begin{cases} \dfrac{u - t_j}{t_{j+2} - t_j} \dfrac{u - t_j}{t_{j+1} - t_j}, & t_j \leq u < t_{j+1}, \\ \dfrac{u - t_j}{t_{j+2} - t_j} \dfrac{t_{j+2} - u}{t_{j+2} - t_{j+1}} + \dfrac{t_{j+3} - u}{t_{j+3} - t_{j+1}} \dfrac{u - t_{j+1}}{t_{j+2} - t_{j+1}}, & t_{j+1} \leq u < t_{j+2}, \\ \dfrac{t_{j+3} - u}{t_{j+3} - t_{j+1}} \dfrac{t_{j+3} - u}{t_{j+3} - t_{j+2}}, & t_{j+2} \leq u < t_{j+3}, \\ 0, & otherwise. \end{cases} \quad (4)$$

If Eq. (3) is expressed as a determinant, the control point can be obtained by converting it to a determinant for the control point.

$$\begin{aligned} [P]_{(m+1)\times 1} &= [N]_{(m+1)\times(n+1)}[Q]_{(n+1)\times 1} \\ [P] &= [N][Q] \\ [N]^T[P] &= [N]^T[N][Q] \\ [Q] &= [[N]^T[N]]^{-1}[N]^T[P] \end{aligned} \quad (5)$$

In the experiment, using Eq. (5), the control points were extracted from the coordinate points of the quadratic B-spline curve that can express the hull form before they were used as the label of the output data. Control points were accurately estimated from the pressure distribution using deep learning, and subsequently, the offsets were extracted through post-processing. Fig. 5 compares the ground-truth stern shape and the stern shape restored by removing the straight section and extracting control points using B-spline.

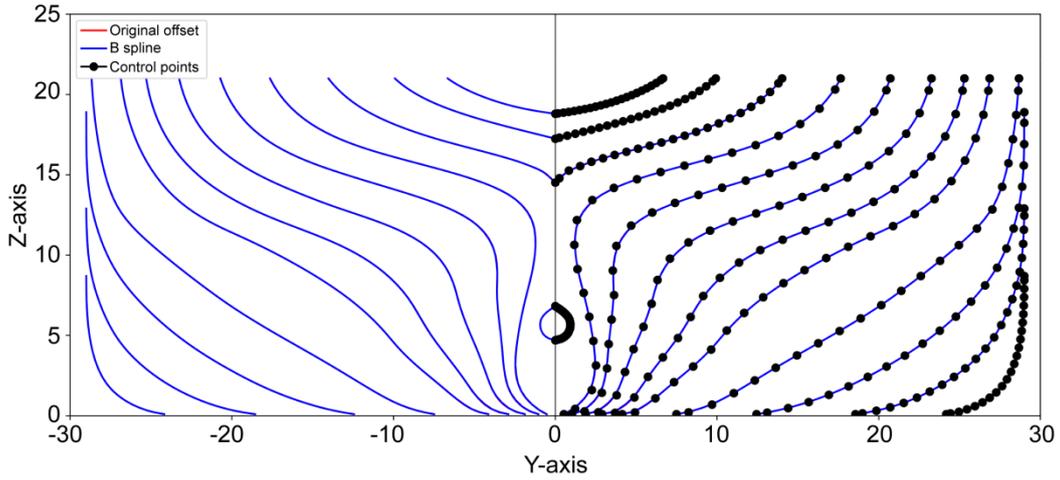

Fig. 5. Comparison between B-spline curve and ground truth curve at 14 sections and 23 control points.

## 4. Multi-task Learning Model

The sections influenced each other; however, because each section has a different shape, errors were generated without convergence when sections were stacked in one pressure distribution image. Therefore, it is necessary to establish a model that can extract control points of multiple sections from one pressure distribution image. This paper proposes a multi-task learning model that can extract shared features from one image and learn and extract each section. As shown in Fig. 6(b), a shared convolutional layer that can extract image features was used as a feature extractor, and the models were compared by changing the shared convolutional layer and composition of the tasks.

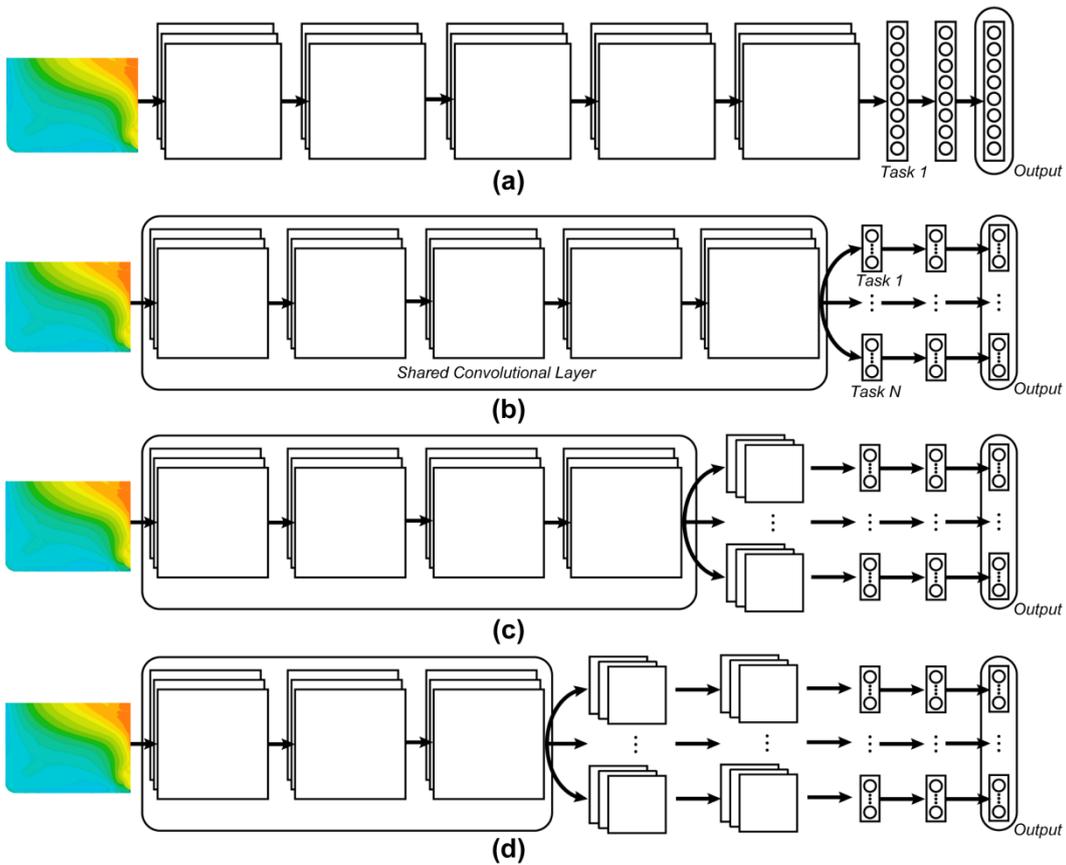

Fig. 6. Convolutional Neural Networks: (a) Single-task learning model, (b) Multi-task learning model (conv0 fc3), (c) Multi-task learning model (conv4 fc3), (d) Multi-task learning model (conv8 fc3). "conv" means the number of convolutional layers included in the task, and "fc" means the number of fully connected layers.

Multi-task learning is designed to optimize multiple objects and a change of the loss function is generated. In the case of general single-task learning, the loss function is defined as follows (Eq. (6)) and learning is performed in the direction of minimizing the loss.

$$\mathcal{L}_{Total\_single} = \sum_{i=1}^{N} \mathcal{L}_i = \sum_{i=1}^{N} f_{loss}(\mathbf{y}_i, F(\mathbf{W}, \mathbf{x}_i)) \qquad (6)$$

$\mathcal{L}$ is the total loss of $N$ learned data. $\mathbf{y}_i = \{q_j\}^{1 \times A}$ denotes a label matrix, $q_j \in \mathbb{R}$ has two data, z-axis, and y-axis values of control point $(n+1)$, which correspond to the width and height of the ship. Because the stern shape is composed of multiple section $S$, it has the values corresponding to $j = 1, \dots, A, A = 2 \times (n+1) \times S$. $\mathbf{W}$ denotes a set of parameters that can be adjusted in the model. $\mathbf{x}_i \in \mathbb{R}^{1 \times D}$ are the features of the input patterns, and $D$ denotes the dimension of the input. $F(\mathbf{W}, \mathbf{x}_i) \in \mathbb{R}^{1 \times A}$, which shows the result of the model, represents CNNs functions such as convolutional layer, pooling layer, and fully connected layer depending on the model configuration. The value estimated by the model $\hat{\mathbf{y}}$ and the label value $\mathbf{y}$ are calculated using a loss function. The same loss function was used from single-task learning and multi-task learning.

$$f_{loss} = \frac{1}{A} \sum_{i=1}^{A} (\mathbf{y}_i - \hat{\mathbf{y}}_i)^2 \qquad (7)$$

For regression of control points, the mean square error (MSE) was used as the loss function (Eq. (7)). The mean is determined by comparing $A$ output data. Then $\mathcal{L}_{Total}$ is calculated through $f_{loss}$, and the parameters of the model are updated through the optimization function. Because multi-task learning determines loss for each task, $\mathcal{L}_{Total}$ is defined as follows:

$$\mathcal{L}_{Total\_multi} = \frac{1}{S} \sum_{i=1}^{N} \sum_{k=1}^{S} \mathcal{L}_{ik} = \frac{1}{S} \sum_{i=1}^{N} \sum_{k=1}^{S} f_{loss}(\mathbf{y}_{ik}, F(\mathbf{W}, \mathbf{x}_i)_k) \qquad (8)$$

Each loss was determined by separating the task for multiple Section $S$ in single-task learning, and the mean loss of the task was calculated. The parameter $\mathbf{W}$ was updated as follows using Adam optimization [32] to minimize the loss.

---

**Algorithm 1** Multi-task learning for Inverse Design

**Input:** Image-Control points pairs $(\mathbf{x}_i, \mathbf{y}_i), i = 1, \dots, N$, and training step $t$.
**Output:** CNNs model parameters $\mathbf{W}$, Predicted control points $\hat{\mathbf{y}}$.
**Loss function:**
    $\mathcal{L}_{Total\_multi}$
**Optimization: Adam**
    **Require:** $\alpha$: Step size
    **Require:** $\beta_1, \beta_2 \in [0,1)$
    **Require:** $\mathcal{L}_{Total}$
    $m_0 \coloneqq 0$ (Initializer initial 1st moment vector)
    $v_0 \coloneqq 0$ (Initializer initial 2nd moment vector)
    $t \coloneqq 0$ (Initializer timestep)
    **While** $\mathbf{W}_t$ not converged **do**
        $t \coloneqq t + 1$
        $g_t \coloneqq \nabla_w f_t(\mathbf{W}_{t-1})$
        $lr_t \coloneqq lr_{t-1} \times \sqrt{1 - \beta_2^t} / (1 - \beta_1^t)$
        $m_t \coloneqq \beta_1 \times m_{t-1} + (1 - \beta_1) \times g_t$

$$v_t := \beta_2 \times v_{t-1} + (1-\beta_2) \times g_t^2$$
$$\mathbf{W} := \mathbf{W}_{t-1} - lr_t \times m_t/(\sqrt{v_t}+\epsilon)$$
**End while**
**Return** $\mathbf{W}_t$ (Resulting parameters)

Starting with the initial learning rate of 0.0001, it was decreased by 1/10 for every 100 epochs to verify loss convergence. The batch size was set to 1, so that one image would be trained in each step.

One single-task learning model and three multi-task learning models were used in this experiment. For the common features of multi-task learning, it has a feature extractor shared by the shared convolutional layers and a regression equation for estimating the control point by section in each task. Multi-task learning improves the inefficient method that divides sections and each section has a single-task learning model. In addition, it enhances the performance of single-task learning that trains every section using one model. The number of parameters is reduced by using shared convolutional layers, and accuracy is improved because section information is reflected by training each section in the task. For the multi-task learning model used in the experiment, some of the shared convolutional layers were distributed to the tasks and its effect on feature extraction was examined. The configuration of each model is shown in Table 1.

Table 1. Model parameters for single-task learning models and multi-task learning models. The white areas are hidden layers included in the shared convolutional layers, and the gray shaded areas are the hidden layers included in the task.

| | | **ConvNet Configuration** | | | | |
|---|---|---|---|---|---|---|
| Layer name | Input feature map | Single Model | Multi-task Model | | | Output feature map |
| | | | Task (conv0 fc3) | Task (conv4 fc3) | Task (conv8 fc3) | |
| Conv1 | 227 × 256 | [3 × 3, 64] × 2 | [3 × 3, 64] × 2 | [3 × 3, 64] × 2 | [3 × 3, 64] × 2 | 113 × 128 |
| Conv2 | 113 × 128 | [3 × 3, 128] × 2 | [3 × 3, 128] × 2 | [3 × 3, 128] × 2 | [3 × 3, 128] × 2 | 56 × 64 |
| Conv3 | 56 × 64 | [3 × 3, 256] × 3 | [3 × 3, 256] × 3 | [3 × 3, 256] × 3 | [3 × 3, 256] × 3 | 28 × 32 |
| Conv4 | 28 × 32 | [3 × 3, 512] × 3 | [3 × 3, 512] × 3 | [3 × 3, 512] × 3 | [[3 × 3, 512] ×3] × 14 | 14 × 16 |
| Conv5 | 14 × 16 | [3 × 3, 512] × 3 | [3 × 3, 512] × 3 | [[3 × 3, 512] ×3] × 14 | [[3 × 3, 512] ×3] × 14 | 7 × 8 |
| FC1 | 7 × 8 | [7 × 8, 4096] | [7 × 8, 512] × 14 | [7 × 8, 512] × 14 | [7 × 8, 512] × 14 | 1 × 1 |
| FC2 | 1 × 1 | [1 × 1, 1000] | [1 × 1, 512] × 14 | [1 × 1, 512] × 14 | [1 × 1, 512] × 14 | 1 × 1 |
| FC3 | 1 × 1 | [1 × 1, 644] | [1 × 1, 46] × 14 | [1 × 1, 46] × 14 | [1 × 1, 46] × 14 | 1 × 1 |
| Output | | | | | | 644 |

The repetition of a convolutional layer and a max pooling layer was set as one Conv. [3×3, 64] indicates the size (3×3) and number of filters (64). The number of filters was increased from 64 to 512. Furthermore, stride is fixed to 1, and the sizes of the input and output feature maps were maintained through padding. After repeating the convolutional layer, a max pooling layer of a 2×2 filter size was placed and down-sampled, and then delivered to the next Conv. FC denotes fully connected layer. Overfitting was prevented by setting dropout for each layer in the training step. The number of sections is 14, and the number of control points in each section is 23 ($n = 22$). The nodes of the FC layer were set considering the difference in the number of outputs of each task between single-task learning and multi-task learning. The information about task in the multi-task learning model is represented by the number of conv and FC. Task (conv0 fc3) indicates that there are three FC layers in the task. Task (conv4 fc3) and Task (conv8 fc3) are models in which the last four/eight shared convolutional layers are placed in each task. Task (conv8 fc3) has more parameters and calculations than the previous model.

# 5. Comparison and Result

The performance of the deep learning model estimating the stern shape corresponding to the pressure distribution was evaluated based on three aspects. The performances were compared according to the image contour of the pressure distribution in the deep learning algorithm, model type composed of multi-task learning, and error when the control point of the B-spline was converted to a coordinate point. Additionally, by applying Grad-CAM [33], the task corresponding to each section was visually confirmed in the affected part of the pressure distribution image. All the experimental results were for the test dataset, and in the model, learning was ended early through an overfitting check.

1. Model performance according to image contour of pressure distribution

By changing the image output type of the pressure distribution based on the multi-task learning model (conv0 fc3), the model performances were compared according to the contour, and the optimal image was used as the reference image for the performance comparison experiment according to the model architecture. Regarding the comparison target image, an image with a contour level of 35 was classified as Case 1, and the image added by the contour line was set as Case 1-1. In this way, an image with a contour level of 25 was created as Case 2 and Case 2-1, and a total of four image types were compared. As shown in Table 2, a result of comparing the control point estimation performance for each case using Root Mean Square Error (RMSE) as an index, a relatively high error occurred in Case 1-1 and Case 2-1 with contour lines. The contour line is black and has a sharply low value compared to the surrounding pixel intensity value. For this reason, Case 1 and Case 2, where the pixel intensity value changes linearly, showed good performance. However, each performance difference is less than 0.1 mm. The error allowed by the actual hull form designer in 3D modeling is less than 10 mm, which is very low when considering the width and height of the ship and the error (design noise) caused by the hull form designer. Although the difference in model performance according to the image type is not large, Case 2, which shows the lowest error among the four cases, was selected as the reference image.

Table 2 Control point RMSE – Task according to the input image for test set (based on conv0 fc4)   Unit: mm

| Image | Sec. 0 | Sec. 1 | Sec. 2 | Sec. 3 | Sec. 4 | Sec. 5 | Sec. 6 | Sec. 7 | Sec. 8 | Sec. 9 | Sec. 10 | Sec. 11 | Sec. 12 | Sec. 13 | Total |
|---|---|---|---|---|---|---|---|---|---|---|---|---|---|---|---|
| Case 1 | 0.513 | 0.501 | 2.206 | 0.193 | 1.663 | 4.392 | 4.141 | 3.655 | 3.187 | 1.905 | 1.293 | 0.997 | 0.817 | 0.712 | **2.227** |
| Case 1-1 | 0.498 | 0.522 | 2.480 | 0.190 | 1.502 | 3.661 | 3.979 | 3.892 | 3.798 | 1.983 | 1.313 | 1.060 | 0.863 | 0.799 | **2.324** |
| Case 2 | 0.353 | 0.348 | 2.465 | 0.141 | 1.416 | 3.198 | 3.461 | 3.264 | 3.123 | 1.677 | 1.017 | 1.057 | 0.513 | 0.509 | **2.008** |
| Case 2-1 | 5.952 | 5.913 | 5.195 | 1.461 | 4.705 | 6.705 | 7.659 | 7.576 | 7.960 | 6.564 | 6.158 | 6.865 | 8.227 | 8.785 | **6.560** |

2. Model performance according to the model type composed of multi-task learning

Performance evaluation for multi-task learning was performed based on the image in Case 2, showing optimal performance. Table 3 shows the results for the four models shown in Table 1 with the RMSE for the control point as an index. All four models satisfy an error of less than 10 mm, which is an allowable error for an actual linear designer in 3D modeling, and the multi-task learning model performed better than the single-task learning model. In particular, Task (conv8 fc3) including the convolutional layer inside the task had the lowest error of 1.684 mm. The error of section increased as the number of inflection points increased, and it is most conspicuous in sections 5~8, which is the stern bulb. Section 0, 1, which corresponds to the ship transom shape, and Section 12, 13, which corresponds to the midship section, have a narrow width of hull form variation and a monotonous curve, so the error appears as small as less than 1 mm. In the case of the single-task learning model, the difference between the average error of sections 5~8 and the average of sections 0, 1, 12, and 13 is 5.21 mm, indicating a higher error than the multi-task learning model. This shows that multi-task learning, a method of learning by calculating the loss of each task, rather than learning by averaging over all points using the existing loss function as in Equation 6, reflects the characteristics of the section.

Table 3 Control point RMSE for test set – Based on input image Case 2   Unit: mm

| Model | Sec. 0 | Sec. 1 | Sec. 2 | Sec. 3 | Sec. 4 | Sec. 5 | Sec. 6 | Sec. 7 | Sec. 8 | Sec. 9 | Sec. 10 | Sec. 11 | Sec. 12 | Sec. 13 | Total |
|---|---|---|---|---|---|---|---|---|---|---|---|---|---|---|---|
| Single-task | 0.607 | 0.601 | 2.778 | 0.378 | 1.990 | 5.435 | 6.234 | 6.076 | 5.880 | 2.632 | 1.611 | 1.448 | 0.844 | 0.742 | **3.840** |

| Task (conv0 fc3) | 0.353 | 0.348 | 2.465 | 0.141 | 1.416 | 3.198 | 3.461 | 3.264 | 3.123 | 1.677 | 1.017 | 1.057 | 0.513 | 0.509 | **2.008** |
| --- | --- | --- | --- | --- | --- | --- | --- | --- | --- | --- | --- | --- | --- | --- | --- |
| Task (conv4 fc3) | 0.207 | 0.149 | 2.856 | 0.191 | 2.155 | 2.479 | 2.831 | 2.730 | 3.019 | 2.423 | 2.354 | 1.607 | 1.221 | 0.391 | **2.057** |
| Task (conv8 fc3) | 0.128 | 0.100 | 2.368 | 0.163 | 1.298 | 2.313 | 2.585 | 2.474 | 2.196 | 2.282 | 1.436 | 1.199 | 0.727 | 0.381 | **1.684** |

3. Conversion from control point to coordinate point

In this section, we evaluated whether the B-spline accurately expresses the original stern shape and confirmed the accuracy of converting the control points estimated by the model into coordinate points through post-processing. As shown in Fig. 7, the RMSE was calculated by dividing the z-axis into 50 equal parts for each section and calculating the corresponding y-coordinate value through interpolation because the original coordinate point and the approximate coordinate point were different. The result when the coordinate points are approximated again using the control point of the b-spline extracted through Equation 5 is the B-spline model in Table 4. In the process of converting the coordinate values using B-spline, an error of 2.946 mm occurs. The error when converting the control point estimated by the model to the coordinate point differs by 2.267 mm for the single-task and 2.23 mm on average for the multi-task learning model. As shown in Fig. 8, it satisfies all the allowable errors of less than 10 mm in 3D modeling, but it is necessary to improve the errors that occur owing to the use of B-spline.

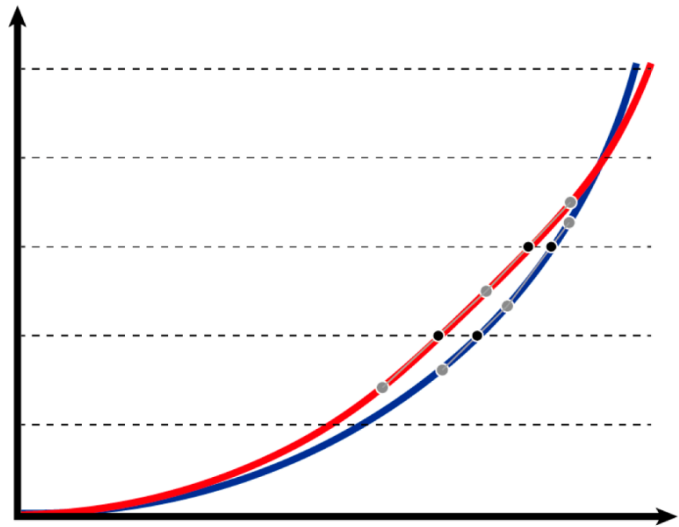

Fig. 7. Interpolation of the original offset and the estimated coordinate point. Interpolation was performed to compare the same points in different sets of comparison coordinate points, and the criterion was set as an interval divided into 50 equal parts on the z-axis.

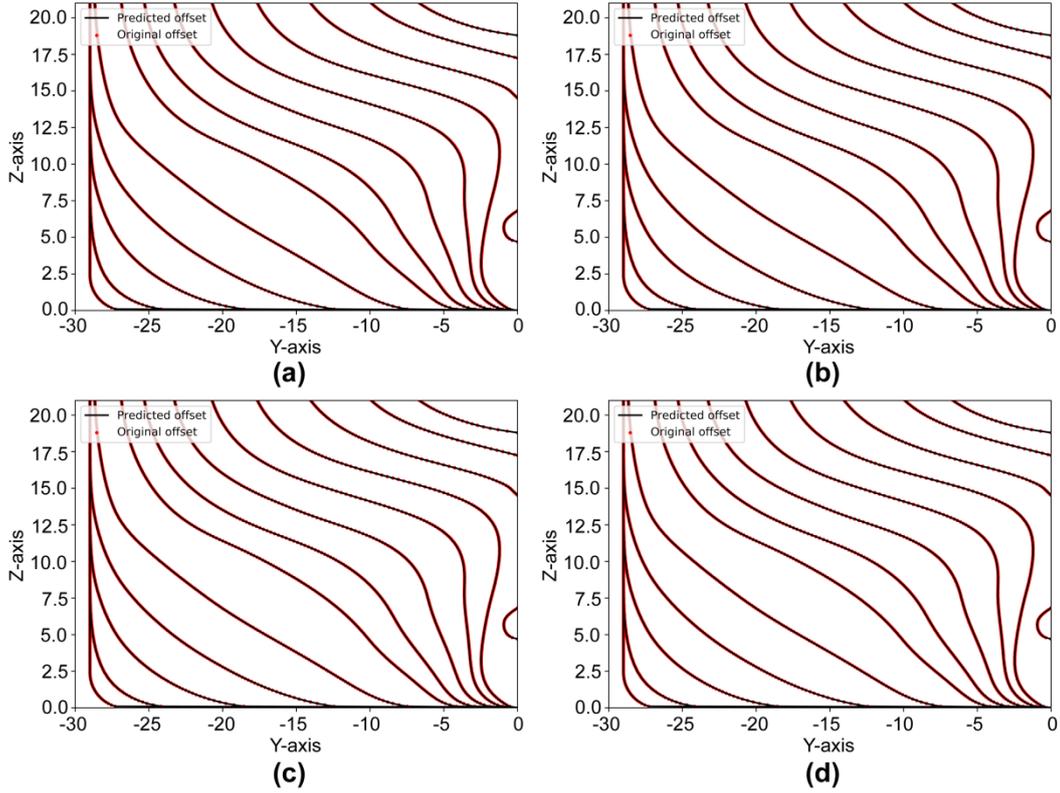

Fig. 8. Ground truth offset & predicted offset
(a) Single-task learning model, (b) Multi-task learning model (conv0 fc3),
(c) Multi-task learning model (conv4 fc3), (d) Multi-task learning model (conv8 fc3)

Table 4 Offset RMSE for test set – Based on input image Case 2  Unit: mm

| Model | Sec. 0 | Sec. 1 | Sec. 2 | Sec. 3 | Sec. 4 | Sec. 5 | Sec. 6 | Sec. 7 | Sec. 8 | Sec. 9 | Sec. 10 | Sec. 11 | Sec. 12 | Sec. 13 | Total |
|---|---|---|---|---|---|---|---|---|---|---|---|---|---|---|---|
| B-spline | 1.496 | 0.615 | 5.710 | 1.031 | 3.573 | 3.815 | 3.183 | 3.355 | 3.499 | 1.711 | 2.422 | 2.365 | 2.790 | 1.456 | **2.946** |
| Single-task | 2.660 | 1.801 | 8.570 | 1.184 | 5.187 | 8.059 | 8.800 | 9.344 | 9.773 | 6.394 | 3.968 | 3.397 | 3.036 | 1.995 | **6.107** |
| Task (conv0 fc3) | 2.196 | 1.240 | 7.748 | 1.108 | 4.183 | 5.636 | 5.165 | 5.283 | 5.687 | 3.669 | 2.901 | 2.884 | 2.886 | 1.627 | **4.183** |
| Task (conv4 fc3) | 1.300 | 0.835 | 8.681 | 1.076 | 4.756 | 4.854 | 4.772 | 4.950 | 5.862 | 4.564 | 3.953 | 3.431 | 3.409 | 1.376 | **4.379** |
| Task (conv8 fc3) | 1.900 | 0.744 | 7.504 | 1.116 | 3.819 | 4.996 | 4.654 | 4.651 | 4.789 | 3.624 | 3.169 | 3.035 | 2.910 | 1.363 | **3.875** |

Because deep learning has a black box structure due to backpropagation, it is not possible to explain in detail how the network results are determined; therefore, it is necessary to check whether each task reflects the actual physical spatial information. Grad-CAM [33] was used to visually check whether each task was related to the stern position affected by the actual section in the trained model. Grad-CAM calculates the gradient and can represent the effect on the result as a heat-map image. Fig. 9 presents the results of using Grad-CAM based on the multi-task learning model (conv0 fc3). In the pressure distribution image of the same size obtained by imaging the estimated coordinate points, sections 0 and 12 tend to show relatively strong intensity, whereas Section 8 tends to show relatively weak intensity; most of the data have significant influence near the section location. The multi-task learning model shares the features extracted through the convolutional layer and estimates the control point corresponding to the section in each task, we can see that it is estimated using the feature corresponding to the section location among the shared features.

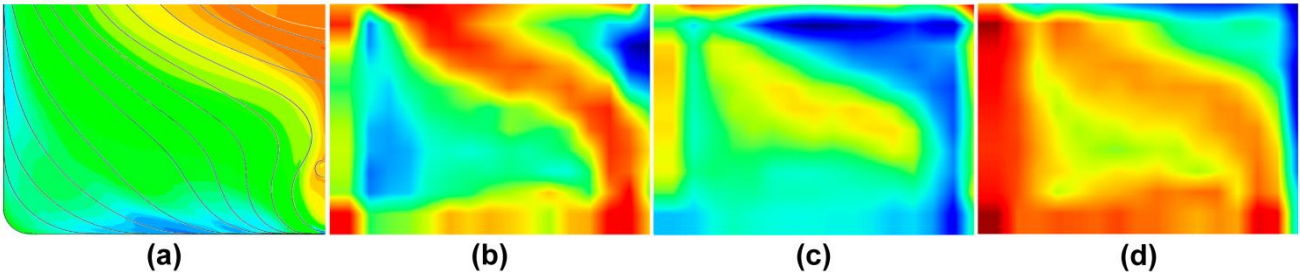

Fig. 9. Result from Grad-CAM: Heat-map image generated based on Case 2, indicates the extent to which sections are affected in the original image. (a) Original image (b) section 0 (c) section 8 (d) section 12

## 6. Conclusion and Future Research

Herein, we proposed an algorithm to inversely design the stern shape that satisfies the pressure distribution using CNNs from deep learning. To fix the number of parameters of the CNNs, the stern shape was used by converting the original offset with a different number for each data (1,002 stern shapes and pressure distribution images) into a fixed number of control points using B-spline. Furthermore, to reflect its various sections, the stern shape was estimated using a multi-task learning model rather than a single-task learning model. The performance according to the wake field flow image type, the performance according to the multi-task learning models, and the performance when the estimated control point is converted into an offset were compared through experiments. Hence, when extracting the stern shape, noise was generated because of the difference between the part where the contour line disappeared and the part where the line was maintained. The data constructed through hull form variation had a variable distribution in sections 5~8 and error of section increased as the number of inflection points increased. Therefore, intensive training is required for the section with many inflection points. Multi-task learning can learn information section-by-section, so the error is 2.156 mm lower than a single-task learning model. An error of 2.946 mm occurred when the hull shape was expressed as a B-spline, and the average error in mm when the estimated control point was converted to an offset was approximately 2 mm, depending on the degree of convergence of the model during the training process.

In future research, we will select appropriate control points and improve the B-spline curve to reduce errors and build data that can flexibly respond to more hull shapes and various hull shapes. In addition, we plan to conduct research to improve deep learning models by improving optimization methods and loss function calculations in multi-task learning.

## Acknowledgement

This study was supported by the research work, 'A Study on the Derivation of Optimum Stern Shape by Analysis of Wake Flow on Ship Based on Deep Learning (I)' funded by the Samsung Heavy Industries co., Ltd.. This work was also supported by the Korea Evaluation Institute Of Industrial Technology(KEIT) and the Ministry of Trade, Industry & Energy(MOTIE) of the Republic of Korea (No. 20018667, Development of Integrated Ship Design System for Hull, Compartment, Basic Calculation and Loading Guidance Based on Artificial Intelligent Technology).## References

[1] Roh, M.-I. and K.-Y. Lee, 2018. Hull Form Design. Computational Ship Design: 141-180.

[2] Huang, T. T., et al., 1976. Propeller/Stern/Boundary-layer interaction on axisymmetric bodies: Theory and Experiment, David W Taylor Naval Ship Research and Development Center Bethesda, MD.

[3] Creutz, G., 1977. Curve and surface design from form parameters by means of B-splines. Ph. D. Thesis, Technical University Berlin.